\documentclass{article}
\usepackage{arxiv}

\usepackage[utf8]{inputenc}
\usepackage[T1]{fontenc}
\usepackage{hyperref}
\usepackage{url}
\usepackage{booktabs}
\usepackage{graphicx}
\usepackage{amsmath}
\usepackage{amsfonts}
\usepackage{nicefrac}
\usepackage{microtype}
\usepackage{cleveref}
\usepackage{natbib}
\usepackage{doi}
\usepackage{xcolor}
\usepackage{tikz}
\usetikzlibrary{shapes, fit, backgrounds, arrows.meta}

\usepackage{listings}
\definecolor{darkblue}{rgb}{0, 0, 0.5}

\title{Skill-based Agentic Evaluation for Real-time Data Science Tasks}

\author{
  \textbf{Aniruddha Tamhane, Raghavendra Addanki, Ayushi Aggarwal, Aditya Bansal,} \\
  \textbf{Rui Wang, Charles Menguy, Swati Jain} \\
  Adobe \\
  \texttt{atamhane@adobe.com}
}
\date{}

\renewcommand{\shorttitle}{Skill-based Agentic Evaluation for Data Science Tasks}

\hypersetup{
  colorlinks=true, citecolor=darkblue, linkcolor=darkblue, urlcolor=darkblue,
  pdftitle={Skill-based Agentic Evaluation for Real-time Data Science Tasks},
  pdfsubject={cs.AI, cs.LG},
  pdfauthor={Aniruddha Tamhane, Raghavendra Addanki, Ayushi Aggarwal, Aditya Bansal, Rui Wang, Charles Menguy, Swati Jain},
  pdfkeywords={agentic evaluation, data science agents, LLM-as-a-judge, ground-truth-as-code, factoid scoring},
}

\begin{document}

\maketitle

\begin{abstract}
We present a framework for evaluating data-science agents on live, continuously updated data using executable ground truth and format-agnostic factoid scoring. Consider this example query: ``what were last week's audience sizes''---the reference answer changes as the underlying data changes, so static references become outdated and standard LLM-as-a-judge pipelines cannot verify responses against a fixed ground truth. Our central contribution, \emph{ground-truth-as-code}, encodes each expected answer as an executable reference function that recomputes the answer directly from live data at evaluation time, ensuring the reference remains consistent with the system it describes. We combine this with a factoid-level, format-agnostic judge that decomposes both the agent's response and the computed ground truth into atomic claims and scores precision, recall, and accuracy over them, irrespective of the response format (prose, list, table, HTML, etc.). The approach is applicable to agents whose expected outputs can be expressed as executable data computations. We validate the framework through a human--LLM agreement study on an internally developed machine learning skill deployed in production, using a synthetic database constructed to reproduce production schemas and entity relationships. Relative to a natural-language ground-truth baseline, our method achieves a $\mathbf{29\%}$ improvement in the Matthews Correlation Coefficient (MCC)---a class-balanced measure of agreement between expert annotators and LLM-as-a-judge predictions---and a $\mathbf{16\%}$ reduction in token consumption per test case, while a self-directed baseline lacking explicit ground truth is anti-correlated with human judgment. Agents that perform multi-source data integration and computation over non-stationary data are routinely deployed in industry; we propose ground-truth-as-code as a practical methodology for their evaluation.
\end{abstract}

\section{Introduction}
\label{sec:intro}

We present a framework for evaluating data-science agents whose outputs depend on live, continuously updated data, combining executable ground truth with format-agnostic factoid scoring. We motivate this design with a concrete example.

Consider a marketing analyst who queries an agent for last week's best-performing campaign segments.
The correct answer depends on the current state of the data warehouse---which rows have been ingested, which schema is active, which tables must be joined---rather than on the answer that was correct at benchmark-authoring time.
A frozen reference answer is therefore stale by construction: by the time it is annotated, the underlying data has already changed the correct response.

Agentic AI systems are increasingly entrusted with precisely this class of deep data computation tasks---multi-source integration, multi-step analysis, and reasoning over large, continuously updated databases---and now underpin production data science workflows across industry \citep{nam2025ds, sun2025survey, chen2025large}.
Evaluating such agents is qualitatively harder than standard LLM evaluation: the property that makes them valuable, operating over live and changing organizational data, violates the time-invariance assumption underlying existing benchmarks and LLM-as-a-judge pipelines, leaving the absence of time-varying ground truth a primary open challenge \citep{chen2026measuring} (Section~\ref{sec:related}).

We address this gap with two ideas, developed in turn below.
First, since the reference answer to a live-data query is fundamentally a computation rather than a static fact, we encode it as such: each expected answer is expressed as a runnable Python function---\emph{ground-truth-as-code}---executed against the live system at evaluation time to recompute the answer from first principles (Section~\ref{sec:impl:gtcode}).
Because the reference is recomputed from the same data state observed by the agent, it remains invariant to data drift, and the function signature serves as a schema contract that fails explicitly when an upstream API changes.

Second, an agent may report identical underlying facts as prose, a table, or a bulleted list, and a single holistic LLM-as-a-judge score conflates presentation quality with the correctness of the underlying data.
We instead have an LLM-based judge decompose both the agent's response and the computed ground truth into atomic claims (\emph{factoids}) and score precision, recall, and accuracy over the matched claims (Section~\ref{sec:impl:metrics}); because the scoring layer never observes the raw response format, the measure is format-agnostic by construction, and hallucinated or omitted claims are directly attributable in the scorecard.
The complete pipeline is implemented as a harness skill, inheriting the harness's native engineering features (Section~\ref{sec:impl}).

We focus on data-science tasks for which the reference answer can be computed from accessible data APIs; extending the approach to settings without such APIs remains outside our current scope.

We validate the framework through a human--LLM agreement study (Section~\ref{sec:human}) on an internally developed, in-production ML skill, using a synthetic database that reproduces the schema, entity relationships, and advancing temporal structure of production data (Section~\ref{sec:synthetic}).
Relative to a natural-language ground-truth baseline, ground-truth-as-code improves the Matthews Correlation Coefficient (MCC)---which quantifies, robustly to class imbalance, how closely the LLM judge's PASS/FAIL verdicts align with expert labels---by $29\%$ ($0.427$ vs.\ $0.331$) and reduces token cost per test case by $16\%$, while a self-directed baseline lacking explicit ground truth is anti-correlated with human judgment ($\mathrm{MCC} = -0.379$).

We contribute: (i) ground-truth-as-code, a technique for constructing drift-resistant reference answers for evaluation on live data; (ii) a factoid-level, format-agnostic scoring framework that separates factual correctness from presentation; and (iii) a validation study against expert judgments on a production-deployed ML skill, showing higher human agreement and lower token cost than natural-language and self-directed baselines.

\section{Related Work}
\label{sec:related}

\textbf{Data Science AI Agents.}
Transformer-based \citep{vaswani2017attention} Large Language Models \citep{brown2020language, white2024livebench} mimic human verbal abilities including conversation, step-by-step reasoning \citep{wei2022chain}, and planning \citep{yao2022react}.
AI agents extend this paradigm by granting LLMs agency through tools, APIs, and MCPs \citep{hou2025model}, and through static or dynamic networks of sub-agents.
Agentic harnesses \citep{pan2026natural} consolidate these capabilities into a standardized substrate that can be specialized via skills \citep{jiang2026sok}---bundles of markdown, scripts, connectors, and tools that augment the harness while reusing its native engineering features (sub-agent orchestration, memory and context management, etc.).
Within this paradigm, production data science workloads span data pulling, aggregation, visualization, and forecasting through to extensive reports and proactive recommendations requiring deep analysis planning, model training, inference, and summarization \citep{nam2025ds, sun2025survey, chen2025large}.
Narrower workflows with strong consistency requirements adopt rigid agentic patterns, with LLMs acting as routers \citep{guo2024ds} or executors of a high-level plan \citep{hong2025data}; embedding planning in the orchestration loop enables open-ended task solving \citep{fu2025autonomous, sun2025data}, and advanced systems such as \citet{nam2025ds} explore multiple concurrent analyses, each with independent planning.
\citet{chen2026measuring} provide a concurrent survey of evaluation tools for data science AI assistants, identifying the lack of time-varying ground truth as a primary open challenge.

\textbf{Agent Evaluation.}
Evaluating advanced agentic capabilities requires test-benches of complex tasks annotated for correctness.
GAIA \citep{mialon2023gaia} pioneered this with a corpus of real-world question--answer pairs demanding complex planning, execution, and information processing.
Data-driven exploration and analysis tasks were curated by \citet{nie2026dsgym, lai2025kramabench}, while \citet{egg2025dabstep} targeted multi-turn problem solving with a factoid-level evaluation setting.
\citet{chen2026dsaeval} evaluate data science agents across a broad range of open-ended real-world tasks, explicitly noting that exact-match and string-overlap metrics are insufficient for such settings.
These works informed our test-bench design; their common limitation is the absence of time- or environment-dependent ground truth.

\textbf{Execution-Based Ground Truth.}
A foundational precedent for our ground-truth-as-code technique is the execution accuracy paradigm established in text-to-SQL evaluation.
WikiSQL \citep{zhong2017wikisql} was among the first to evaluate NL-to-code by executing generated SQL and comparing outputs rather than matching query strings, establishing that code correctness is best measured through its observable effect.
Spider \citep{yu2018spider} scaled this across 200 complex databases, and BIRD \citep{li2023bird} extended it to real-world enterprise settings requiring external knowledge reasoning.
Spider 2.0 \citep{lei2024spider2} pushes further into enterprise agentic workflows where even frontier models solve fewer than one in five tasks.
We transfer this execution-accuracy philosophy to open-ended data science agent responses on live production data, generalizing beyond SQL to arbitrary Python computations.

\textbf{LLM-as-a-Judge and Atomic Factoid Evaluation.}
Agentic responses are inherently unstructured, motivating LLM-as-a-judge approaches \citep{gu2024survey, zheng2023judging}, though these have known limitations on specialized content \citep{szymanski2025limitations}.
Agents-as-Judge \citep{zhuge2024agent} extends the idea by granting the judge limited interaction capabilities.
Factoid-level evaluation---decomposing free-text into atomic claims and scoring each independently---was established by FActScore \citep{min2023factscore}, which achieves near-human factual precision measurement on long-form generation.
SAFE \citep{wei2024longfact} extends this by deploying an LLM sub-agent to verify each atomic fact via search, introducing sub-agent parallelism as a natural implementation pattern.
\citet{jafari2026beyond} and \citet{fan2025minoseval} target factoid-level precision and recall directly; our framework extends these ideas with conditioning on the user query when extracting ground-truth factoids (Eq.~\ref{eq:factoids}).
Retrieval-augmented generation (RAG) evaluation frameworks occupy an adjacent but architecturally distinct niche.
RAGAS \citep{es2023ragas} decomposes RAG pipeline quality into four LLM-judged metrics---Faithfulness, Answer Relevancy, Context Precision, and Context Recall---each requiring the retrieved context chunks as an explicit input; the framework is thus inapplicable to agents that produce answers through API calls or code execution rather than document retrieval.
ARES \citep{saadfalcon2023ares} similarly targets RAG pipelines, fine-tuning lightweight LLM judges on synthetic preference data to score Context Relevance, Answer Faithfulness, and Answer Relevance, using prediction-powered inference for statistical confidence intervals.
Both systems assume a static document corpus and have no mechanism for ground truth that must be re-executed against a live data system.
Our framework extends the factoid-decomposition spirit of these approaches to the agentic data science setting, replacing retrieved-context fidelity with executable-code ground truth evaluated against production data at inference time.

\textbf{Dynamic and Contamination-Free Evaluation.}
Static benchmarks face two compounding failure modes.
First, ground truth derived from world-state facts degrades as production data drifts: \citet{margatina2023dynamic} demonstrate this with temporal concept drift in language model benchmarks, and \citet{shi2025benchmarks} quantify the resulting score inflation for static factuality test sets.
Second, training-data contamination of widely-shared benchmarks systematically overstates model capability \citep{xu2025contamination}.
Live-data evaluation frameworks such as LiveBench \citep{white2024livebench} and PolyBench \citep{arora2026polybench} partially address contamination by continuously refreshing questions from real-world streams, but retain fixed answer verification logic that cannot accommodate changing production schemas.
Our ground-truth-as-code approach addresses both failure modes: the Python function is re-executed at evaluation time against live data, making both the answer and its verification inherently drift-resistant.

\section{Implementation}
\label{sec:impl}

\subsection{Framework Overview}
\label{sec:impl:overview}

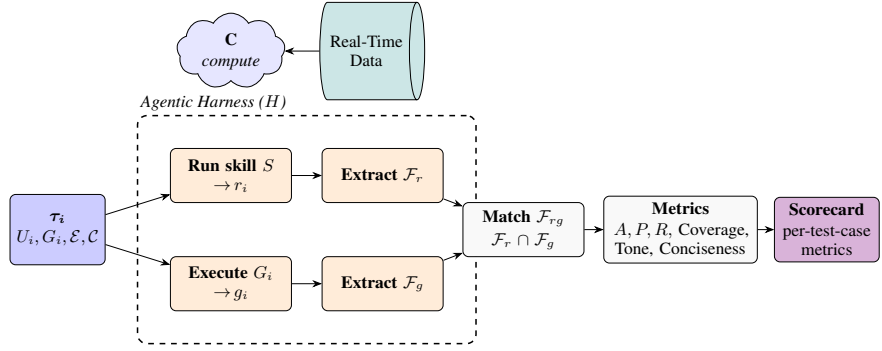
\begin{figure*}[t]
\centering
\resizebox{0.7\textwidth}{!}{%
\begin{tikzpicture}[
    >=Stealth,
    font=\small,
    hbox/.style={draw, rounded corners=3pt, fill=orange!15,
                 minimum height=1.0cm, minimum width=2.2cm,
                 align=center, text width=2.0cm},
    pbox/.style={draw, rounded corners=3pt, fill=gray!5,
                 minimum height=1.0cm, minimum width=2.2cm,
                 align=center, text width=2.0cm},
    tbox/.style={draw, rounded corners=3pt, fill=blue!20,
                 minimum height=1.3cm, minimum width=1.3cm,
                 align=center},
    sbox/.style={draw, rounded corners=3pt, fill=violet!30,
                 minimum height=1.0cm, minimum width=1.8cm,
                 align=center, text width=1.7cm},
]
    \node[tbox] (tau) at (0, 0)
        {$\boldsymbol{\tau_i}$\\[1pt]\footnotesize$U_i, G_i, \mathcal{E}, \mathcal{C}$};

    \node[hbox] (runS) at (3.2, 1.0)
        {\textbf{Run skill $S$}\\${\to}\,r_i$};
    \node[hbox] (extFr) at (6.0, 1.0)
        {\textbf{Extract $\mathcal{F}_r$}};

    \node[hbox] (execG) at (3.2, -1.0)
        {\textbf{Execute $G_i$}\\${\to}\,g_i$};
    \node[hbox] (extFg) at (6.0, -1.0)
        {\textbf{Extract $\mathcal{F}_g$}};

    \node[pbox] (match) at (8.6, 0)
        {\textbf{Match $\mathcal{F}_{rg}$}\\[2pt]\footnotesize$\mathcal{F}_r \cap \mathcal{F}_g$};

    \node[pbox, minimum width=2.8cm, text width=2.6cm] (metrics) at (11.5, 0)
        {\textbf{Metrics}\\[1pt]\footnotesize$A, P, R,$ Coverage, Tone, Conciseness};

    \node[sbox] (scorecard) at (14.2, 0)
        {\textbf{Scorecard}\\[1pt]\footnotesize per-test-case\\metrics};

    \node[cloud, draw, fill=blue!10, cloud puffs=9, cloud puff arc=120,
          minimum width=2.0cm, minimum height=1.4cm] (cloud) at (3.2, 3.3) {};
    \node[align=center] at (cloud.center)
        {\textbf{C}\\[1pt]\footnotesize\textit{compute}};

    \node[cylinder, draw, fill=teal!20, shape aspect=0.35,
          minimum width=1.8cm, minimum height=0.9cm,
          align=center, font=\footnotesize] (db) at (5.7, 3.3)
          {Real-Time\\Data};

    \begin{pgfonlayer}{background}
        \node[draw, dashed, thick, rounded corners=5pt, inner sep=0.6cm,
              fit=(runS)(execG)(extFr)(extFg)] (harness) {};
    \end{pgfonlayer}
    \node[font=\footnotesize\itshape, anchor=south west, inner sep=2pt]
        at (harness.north west) {Agentic Harness ($H$)};

    \draw[->] (tau) -- (runS);
    \draw[->] (tau) -- (execG);
    \draw[->] (runS) -- (extFr);
    \draw[->] (execG) -- (extFg);
    \draw[->] (extFr) -- (match);
    \draw[->] (extFg) -- (match);
    \draw[->] (match) -- (metrics);
    \draw[->] (metrics) -- (scorecard);

    \draw[->] (db) -- (cloud);

\end{tikzpicture}
}%
\caption{Evaluation flow for a single test case $\tau_i = (U_i, G_i, \mathcal{E}, \mathcal{C})$. Both the skill response $r_i$ and the ground-truth output $g_i$ are produced at evaluation time against live data. Factoids are extracted from each, matched semantically, and aggregated into a per-test-case scorecard.}
\label{fig:evaluation-flow}
\end{figure*}

The pipeline assesses factual equivalence between the agent's response and a reference ground truth, conditioned on the user query and on conversational expectations (conciseness, coverage, tone); Figure~\ref{fig:evaluation-flow} shows the full flow.
Given a skill $S$ under evaluation and an agentic harness $H$, we build a test-bench $T = \{\tau_i\}_{i=1}^{t}$ of tuples $\tau_i = (U_i,\, G_i,\, \mathcal{E},\, \mathcal{C})$: the user query $U_i$, the ground-truth-as-code function $G_i$, environment variables $\mathcal{E}$ (business context, co-resident skills), and compute variables $\mathcal{C}$ (credentials, timeouts, API endpoints).
Crucially, the reference ground truth is never precomputed or stored: for each $\tau_i$ the harness runs $S$ and executes $G_i$ \emph{live at evaluation time}, in parallel and against the same current state of the data system, to collect the response $r_i$ and the freshly computed ground-truth output $g_i$, then compares them at the level of atomic factual claims.

\subsection{Ground-Truth-as-Code}
\label{sec:impl:gtcode}

Each test case encodes its expected answer not as a static string but as a typed Python function $G_i: \mathcal{E} \rightarrow \mathcal{D}$, where $\mathcal{D}$ maps factoid keys to their expected values.
At evaluation time, $G_i$ is executed within the compute environment specified by $\mathcal{C}$, which connects to the live data system and pulls real-time data from the cloud to produce $g_i = G_i(\mathcal{E})$.
Because $g_i$ is computed from the same current data state as the response $r_i$, the reference resists data drift; and being ordinary code, $G_i$ is versioned with the test-bench in source control, giving the same per-change auditability as production code.

\subsection{Factoid Decomposition}
\label{sec:impl:factoids}

A \emph{factoid} is an atomic claim---the smallest unit of information independently verifiable as true or false.
Let $\Phi(\cdot)$ map a string to its set of atomic claims; we apply it asymmetrically to the response and ground-truth:
\begin{equation}
    \mathcal{F}_r = \Phi(r_i), \qquad \mathcal{F}_g = \Phi(g_i \mid U_i).
    \label{eq:factoids}
\end{equation}
The response operator is unconditioned---a well-behaved agent surfaces only facts relevant to $U_i$, so any extraneous claim in $r_i$ is itself a scoreable signal---while the ground-truth operator is conditioned on $U_i$, keeping only the factoids in $g_i$ that bear on the query.
Extraction is \emph{format-independent}: regardless of the surface format of $r_i$, $\mathcal{F}_r$ contains the same atomic claims.
This is the source of the framework's style agnosticism---the scoring layer never observes the raw response format, only the extracted factoids.

\subsection{Factoid Matching}
\label{sec:impl:matching}

Matching is performed by the agentic harness, which attempts a one-to-one mapping between the response factoids $\mathcal{F}_r$ and the ground-truth factoids $\mathcal{F}_g$ (the latter already conditioned on the user query $U_i$); $\mathcal{F}_{rg}$ is the resulting set of matched factoids.
Each match is a binary decision (no partial credit), so a matched factoid is unambiguously correct and an unmatched one unambiguously wrong, preserving the interpretability of the metrics below.

\subsection{Metrics}
\label{sec:impl:metrics}

\textbf{Factoid-level metrics.}
We define accuracy ($A$), precision ($P$), and recall ($R$):
\begin{align}
    A &= \frac{|\mathcal{F}_{rg}|}{|\mathcal{F}_g| + |\mathcal{F}_r| - |\mathcal{F}_{rg}|}, \nonumber \\
    P &= \frac{|\mathcal{F}_{rg}|}{|\mathcal{F}_r|}, \qquad
    R = \frac{|\mathcal{F}_{rg}|}{|\mathcal{F}_g|}.
    \label{eq:metrics}
\end{align}
$P$ is the fraction of agent-stated facts that are correct (low $P$ indicates hallucination), $R$ the fraction of expected facts surfaced (low $R$ indicates omission), and $A$ the overall accuracy over factoids.
The metrics are interpretable by construction: each scorecard logs $\mathcal{F}_r \setminus \mathcal{F}_{rg}$ (hallucinated claims) and $\mathcal{F}_g \setminus \mathcal{F}_{rg}$ (omitted facts) verbatim, enabling direct inspection of failure modes.

\textbf{Qualitative dimensions.}
Beyond factual overlap, a rubric-guided LLM judge scores three further dimensions: \textbf{Question Coverage}, how completely the explicit sub-questions or action items in $U_i$ are addressed by $r_i$; \textbf{Tone}, the politeness and professionalism of $r_i$; and \textbf{Conciseness}, the verbosity of $r_i$ relative to the information conveyed.
Each rubric anchors its scale to reduce intra-judge variance across test cases and agent versions.

For consistency across all dimensions, every factoid-level and qualitative metric is reported on a common $0$--$9$ scale, obtained by a simple linear rescaling of the underlying $[0, 1]$ score (i.e., multiplying by $9$).

\section{Synthetic Data Generation}
\label{sec:synthetic}

To validate the framework with known ground truth, we construct a reproducible synthetic database that mirrors the schema, entity relationships, and continuously advancing temporal structure of the production customer data.
It is deployed on a live stage sandbox behind the same APIs the agent and ground-truth functions $G_i$ must query, mirroring the query-then-execute pathway of production rather than a static local mock---so moving to production data requires only a credentials swap.
A Poisson--DAG engine generates user-behavior events, customer profiles, and a domain catalog across three verticals (retail, financial services, healthcare), each over a rolling 26-week window calibrated to the production system's event rates, cohort sizes, and catalog cardinalities (Table~\ref{tab:synthetic-data}).
Heterogeneous domain vocabularies (e.g., ``prescription fill'' vs.\ ``cart abandonment'') stress the ML-goal identification and dataset-discovery sub-tasks of Section~\ref{sec:human}.

\begin{table}[t]
    \centering
    \small
    \caption{Vertical-wise summary statistics for the primary 6-month synthetic benchmark bundles. Each bundle contains experience events (EE), customer/patient profiles, and a domain catalog deployed on the stage sandbox.}
    \begin{tabular}{lrrrr}
        \toprule
        Vertical & EE Records & Profiles & Catalog & Window \\
        \midrule
        Retail               &  17,825 & 10,000 & 50 & 26 wk \\
        Fin.\ Services       &  49,651 & 10,000 & 12 & 26 wk \\
        Healthcare           & 285,854 & 10,000 & 20 & 26 wk \\
        \midrule
        \textbf{Total}       & \textbf{353,330} & \textbf{30,000} & \textbf{82} & \\
        \bottomrule
    \end{tabular}
    \label{tab:synthetic-data}
\end{table}

Profiles use fixed seeds and realistic within-vertical distributions---credit tiers and income bands (financial), insurance types and chronic-condition flags (healthcare), loyalty tiers (retail)---and a realism layer injects column confounds (plausible but semantically misleading column names) to prevent trivially correct schema matches.

\textbf{Decoy datasets.}
Because na\"{i}ve retrieval can succeed by lexical overlap alone, we inject decoys confusable with a correct bundle along a single dimension yet unsuitable for the full ML pipeline: (i) a profile-only snapshot missing the events and catalog needed for training; (ii) an events-only extract that cannot be joined to profiles; (iii) a cross-vertical billing dataset whose event names overlap financial terminology; (iv) a mislabeled legacy dataset claiming a 6-month window but holding only 2 weeks; and (v) a marketing audience export overlapping the retail profile namespace.
We further add 2-week-window bundles across five verticals that share schema with their 6-month counterparts but are too short for the propensity tasks, trapping agents that ignore data recency.
Satisfying a ground-truth function requires identifying \emph{all three} datasets of a bundle---events, profiles, and catalog---so any decoy selection incurs a precision penalty.

\section{Human--LLM Correlation Study}
\label{sec:human}

\textbf{Test-bench.}
We validate the LLM-as-a-judge component on 55 single-turn test cases targeting two sub-tasks of an in-house data science skill---ML-goal identification and dataset discovery from an abstract user request---backed by the synthetic database of Section~\ref{sec:synthetic}.
Each case pairs a user request tied to a product-level business goal (e.g., ``\textit{I'd like a loan application propensity model to predict which customers will seek a personal loan next quarter}'') with ground-truth code encoding $ml\_goal$, $search\_terms$, and $expected\_datasets$; environment and compute variables are inherited from the test skill.
Of the 55 cases, 53 are comparable for the pass/fail analysis.

\textbf{Experimental setup.}
In our experiments the agentic harness is Claude Code and the underlying LLM is Claude Sonnet 4.6, used both by the skill under test and by the LLM-based evaluation operators (e.g., factoid extraction and the rubric-guided qualitative judge).
The framework is nonetheless model- and harness-agnostic: any sufficiently capable LLM and any agentic harness can be substituted without changes to the test-bench or metrics.

\textbf{Skill under test.}
The skill under evaluation is an in-house machine learning skill that parses the abstract user query, discovers the relevant datasets from the live data system, and constructs the input variables and target features for a downstream propensity model---all expressed as executable Python code.

\textbf{Annotation.}
After a calibration round, three domain-expert annotators independently rate each response against the ground truth, with live-data access for fact verification, on (a) a 5-point Likert per atomic factoid and (b) a 5-point Likert across three qualitative dimensions (Coverage, Conciseness, Tone).
The LLM judge emits an execution score $\mathrm{exec\_score} \in [0, 9]$; human ratings are rescaled to the same range via $(x-1)\times 9/4$.

\textbf{Study quality.}
Prevalence-adjusted Gwet's $\mathrm{AC}_2$ clears the $0.667$ acceptability threshold at both the factoid level ($0.931$) and the stacked-dimension level ($0.883$), confirming the annotations are trustworthy (Table~\ref{tab:iaa}).

\begin{table}[t]
\centering
\caption{Inter-annotator agreement (Gwet's $\mathrm{AC}_2$, prevalence-adjusted) at the factoid level and over all dimensions; both clear the $0.667$ acceptability threshold.}
\label{tab:iaa}
\small
\begin{tabular}{lc}
\toprule
\textbf{Level} & \textbf{$\mathrm{AC}_2$} \\
\midrule
All dimensions (overall) & \textbf{0.883} \\
Factoid-level            & \textbf{0.931} \\
\bottomrule
\end{tabular}
\end{table}

\subsection{Baseline Comparison}
\label{sec:human:baseline}

We compare ground-truth-as-code against two baselines, each weakening a different aspect of the reference given to the agentic-harness evaluator (Table~\ref{tab:baseline}).
We report the Matthews Correlation Coefficient (MCC)---a single-figure correlation score, insensitive to skewed class distributions, capturing the alignment between expert annotations and the LLM judge's PASS/FAIL predictions---alongside precision, recall, and F1.

\textbf{Baseline-1 (No GT)} provides no explicit ground truth at all.
Instead, it repurposes the agentic harness's native \emph{skill-creator} skill---a built-in component for authoring and evaluating skills---with a prompt that nudges it to study the skill under test and devise its own criteria for correctness.
This baseline isolates the merit of supplying \emph{any} explicit ground truth to a harness-based LLM evaluator: if a capable harness can reliably infer correctness on its own, an externally provided reference would be redundant.

\textbf{Baseline-2 (NL GT)} supplies the same expected answer as our ground-truth-as-code, but expressed as natural-language instructions rather than runnable code---a textual specification of how to compute the answer and what to expect, which the evaluator must interpret and act on itself.
This baseline isolates the merit of generating the ground truth in a structured, executable form (code) as opposed to an unstructured natural-language set of instructions.

The two baselines underperform ground-truth-as-code in distinct, instructive ways (exact figures in Table~\ref{tab:baseline}).
Baseline-1 consistently evaluated each response only against the final-answer checks explicitly declared as \texttt{final checks} in the skill under test; it never formulated an alternate pathway to compute the ground truth and thus never accessed the real-time data.
Grading the skill against its own asserted behavior leaves its agreement anti-correlated with human judgment, and at the highest token cost, since self-devising a checking plan is token-heavy.
Baseline-2 did invoke the correct API function, but over-specified some of its input parameters; the resulting reference carried extra factoids that occasionally confounded the LLM judge, depressing precision while still incurring the token cost of generating a natural-language reference and invoking the skill.
Ground-truth-as-code instead recomputes the reference deterministically against live data, attaining the best human agreement (highest MCC and F1, full recall) at the lowest token cost.
The most precise and token-efficient ground-truth computation is therefore achieved through the ground-truth-as-code implementation.

\begin{table}[t]
\centering
\caption{Baseline comparison against the three-annotator human reference ($53$ TCs): PASS/FAIL agreement (MCC), precision, recall, F1, and mean per-test-case token cost.}
\label{tab:baseline}
\small
\begin{tabular}{lccccc}
\toprule
\textbf{Configuration} & \textbf{MCC} & \textbf{Prec.} & \textbf{Rec.} & \textbf{F1} & \textbf{Tokens/TC} \\
\midrule
Baseline-1: No GT  & $-$0.379 & 0.238 & 0.200 & 0.217 & ${\sim}$70k \\
Baseline-2: NL GT  & 0.331    & 0.561 & 0.920 & 0.697 & 29.2k \\
\midrule
\textbf{GT-as-code (ours)} & \textbf{0.427} & \textbf{0.568} & \textbf{1.000} & \textbf{0.724} & \textbf{${\sim}$24.6k} \\
\bottomrule
\end{tabular}
\end{table}

\section{Conclusions}
\label{sec:conclusion}

In conclusion, we present a skill-based agentic evaluation framework for data science AI agents operating on real-time data, addressing challenges that static offline benchmarks cannot: temporally dynamic ground truth and format-agnostic fact verification.
Ground-truth-as-code makes the reference drift-resistant by executing against live data at evaluation time.
Style-agnostic factoid-level metrics separate factual correctness from response format, and the per-factoid scoring is interpretable by construction: developers can read off hallucination rates, omission rates, and failure-clustering by fact type directly from the scorecard.
The harness-skill implementation delivers concurrent, logged evaluation that evolves alongside the systems it assesses, and the actionable per-factoid verdicts close the loop from measurement to targeted improvement.
In our study, this executable reference agreed with human annotators more closely than natural-language or self-directed baselines while costing fewer tokens, evidencing the practical payoff of ground-truth-as-code.

\section*{Limitations}
\label{sec:limitations}

We discuss the limitations of our study below and, where applicable, propose mitigations for future work.

\textbf{Self-judging bias.}
In our experiments the same underlying LLM (Claude Sonnet 4.6) powers both the skill under test and the LLM-based evaluation operators (factoid extraction and the rubric-guided qualitative judge), which in principle exposes the evaluation to self-preference bias, in which a judge favors outputs from its own model family.
We argue that this risk is substantially mitigated by our design: the judge does not score free-form responses holistically but is driven by a clearly defined rubric that quantifies correctness against an externally computed, executable ground truth at the level of atomic factoids.
Grounding every verdict in rubric-anchored, factoid-level comparisons against ground-truth-as-code leaves little room for stylistic self-preference to influence the score.
Nonetheless, a systematic study that varies the judge model independently of the skill model remains valuable future work.

\textbf{Completeness of the ground-truth code.}
The correctness guarantees of the framework rest on the assumption that the ground-truth-as-code function $G_i$ generates correct and complete reference data for \emph{all} sub-queries implied by the user request $U_i$.
If $G_i$ omits a relevant sub-query or computes it incorrectly, the resulting factoid-level metrics will misjudge the agent accordingly.
Ensuring the completeness and correctness of the ground-truth code is therefore a precondition for reliable evaluation, and currently rests on expert authoring and review.

\textbf{Annotation effort.}
Curating ground-truth-as-code requires more expert annotation effort than a natural-language reference; a systematic method for auto-generating the ground-truth code remains to be explored and has not been studied in this paper.

\textbf{Higher-capacity models.}
The efficacy of natural-language ground truth should be re-examined with higher-capacity models such as Claude Opus or OpenAI GPT-5.5, which may interpret and explore their computational environments and follow unstructured instructions more reliably.

\textbf{Robustness of the token-efficiency claim.}
The robustness of the token-efficiency claim needs to be tested by running similar experiments on agentic harnesses with different memory and tool-call caching mechanisms.

\section*{Ethics Statement}

\textbf{Annotator participation.}
All annotators took part on a fully voluntary basis. They were either full-time employees or paid interns, completely independent from the study itself, and none had any stake in its outcome; no compensation was tied to the ratings they provided.

\textbf{Data privacy.}
The synthetic database used throughout our evaluation was constructed specifically to prevent leakage of customer privacy of any sort. Absolutely no real-world data was used in either the creation of the synthetic data or the evaluation; all profiles, events, and catalogs are programmatically generated from fixed seeds and contain no personally identifiable information.

\textbf{Broader impact.}
Our framework can potentially be used to evaluate other data-reporting systems on factual accuracy, helping to surface and quantify hallucinated facts. To mitigate AI-generated hallucination of facts, the ground truth is encoded as runnable code rather than free-form text, so the reference is computed deterministically rather than generated. Because this code is executed within a secure, sandboxed compute environment, the risk of malicious code being executed by the AI is contained.

\bibliographystyle{unsrtnat}
\bibliography{references}

\appendix
\section*{Appendix A: Worked Example of a Single Test-Case Evaluation}
\label{app:worked-example}

To make the pipeline of Section~\ref{sec:impl} concrete, we walk through the evaluation of one real test case end to end.
The example is test case $\tau_0$ from the v3 data-discovery benchmark (financial vertical, \texttt{tag = financial-modified}), with components $(U_0, G_0, \mathcal{E}, \mathcal{C})$: the user query $U_0$, the ground-truth-as-code function $G_0$, and the environment and compute variables $\mathcal{E}, \mathcal{C}$ inherited from the skill under test.
All content below is reproduced verbatim from our evaluation artifacts; long text is abridged with ``\texttt{[...]}''.

\subsection{Input query $U_0$}
\begin{quote}
\itshape
Our risk committee requested this after reviewing last quarter's delinquency rates. We need a propensity model to flag credit card accounts showing early signs of default risk---find the relevant AEP datasets, identify the key input variables, the target variable, and define the target mapping function.
\end{quote}

\subsection{Ground-truth-as-code $G_0$}
The reference answer is expressed as an \emph{executable function invocation}, not as literal expected values. The test case stores the call
\begin{lstlisting}[language=Python]
get_ground_truth_string(idx=0, gt_keys=['ml_goal_type', 'target_variable',
    'expected_datasets', 'input_variables', 'target_mapping_function',
    'observation_window_days', 'outcome_window_days', 'failure_modes'])
\end{lstlisting}
which invokes a reference function declared by the test bench.
Executing $G_0$ at evaluation time produces the computed ground truth $g_0$ below. In this benchmark the function retrieves and formats the reference specification for the case; the same code-based interface generalizes to reference functions that recompute the answer from live data (Section~\ref{sec:impl:gtcode}).

\subsection{Computed ground truth $g_0 = G_0(\mathcal{E})$}
The string returned by executing $G_0$ specifies:
\begin{itemize}
\item \textbf{ML goal:} \texttt{binary\_classification}.
\item \textbf{Target variable:} \texttt{default\_risk\_30d}.
\item \textbf{Expected datasets:} the Financial 6M bundle (\texttt{69bb43ac\dots}, \texttt{69bb43ad3\dots}, \texttt{69bb43ad7\dots}).
\item \textbf{Input variables (with roles):} \texttt{creditTier} (feature), \texttt{incomeBand} (confounder), \texttt{accountBalance} (feature); \texttt{count\_90d} aggregations of the events \texttt{account.login}, \texttt{creditCard.purchase}, and \texttt{billing.paymentReceived}; \texttt{riskBand} (confounder); and \texttt{lastPaymentFailedAt} (\emph{leakage}---must be excluded from training).
\item \textbf{Target mapping function:} let $t_0 = \max(\text{timestamp})$ per user; $y=1$ if a \texttt{billing.paymentFailed} event occurs in $(t_0, t_0 + 30\text{d}]$, else $0$; features use events in $(t_0 - 90\text{d}, t_0]$; exclude \texttt{lastPaymentFailedAt}.
\item \textbf{Windows:} observation $90$ days, outcome $30$ days.
\end{itemize}

\subsection{Agent response $r_0$}
The agent's response (abridged to the scored sections) selected the \emph{base} Financial datasets rather than the Financial 6M bundle, and defined the target from a utilization heuristic rather than the \texttt{billing.paymentFailed} event:
\begin{quote}
\ttfamily\footnotesize
\textbf{Datasets Selected:} [Financial] Experience Events \texttt{69b33da5\dots}, [Financial] Customer Profiles \texttt{69b33da6\dots}, [Financial] Account Catalog \texttt{69b33da6\dots}. [...] \\[2pt]
\textbf{Target Variable:} \texttt{is\_default\_risk} (binary). [...] \\[2pt]
\textbf{Target Mapping Function:} is\_default\_risk = 1 if \texttt{utilization\_rate > 0.85} OR \texttt{transactionType == 'missed\_payment'}, else 0. [...]
\end{quote}
Profile features (\texttt{creditTier}, \texttt{incomeBand}, \texttt{accountBalance}) and the binary-classification goal are correct; the response includes no data-leakage warning.

\subsection{Factoid extraction, matching, and metrics}
The judge extracts factoids from the response and the computed ground truth (Eq.~\ref{eq:factoids}), then matches them (Section~\ref{sec:impl:matching}).
For this case the judge counts approximately $|\mathcal{F}_r| \approx 14$ response factoids, $|\mathcal{F}_g| \approx 12$ query-relevant ground-truth factoids, and $|\mathcal{F}_{rg}| \approx 5$ matched factoids.

Matching is one-to-one and binary (Section~\ref{sec:impl:matching}): each response factoid either matches exactly one ground-truth factoid or is left unmatched, with no partial credit. The result partitions the factoids into three groups.
\begin{itemize}
\item \textbf{Matched} ($\mathcal{F}_{rg}$, ${\approx}5$): the binary-classification goal; the input variables \texttt{creditTier}, \texttt{incomeBand}, and \texttt{accountBalance}; and the binary target concept. These are the facts the response and ground truth agree on.
\item \textbf{Ground-truth factoids with no match} ($\mathcal{F}_g \setminus \mathcal{F}_{rg}$, ${\approx}7$) --- omissions that lower recall: the Financial 6M dataset IDs, the \texttt{billing.paymentFailed} target event, the \texttt{count\_90d} event aggregations (\texttt{account.login}, \texttt{creditCard.purchase}, \texttt{billing.paymentReceived}), the $90$-day observation and $30$-day outcome windows with $t_0$ anchoring, and the \texttt{lastPaymentFailedAt} leakage exclusion.
\item \textbf{Response factoids with no match} ($\mathcal{F}_r \setminus \mathcal{F}_{rg}$, ${\approx}9$) --- claims that lower precision: the base Financial dataset IDs (wrong bundle), the target defined from \texttt{utilization\_rate > 0.85}, the value \texttt{transactionType = 'missed\_payment'} (absent from the schema), and additional features proposed but not in the ground truth (\texttt{creditLimit}, \texttt{interestRate}, \texttt{monthlyFee}, \texttt{primaryAccountType}, derived customer age).
\end{itemize}
Substituting the three counts into Eq.~\ref{eq:metrics}:
\begin{equation*}
P = \tfrac{5}{14} \approx 0.36, \qquad
R = \tfrac{5}{12} \approx 0.42, \qquad
A = \tfrac{5}{14 + 12 - 5} = \tfrac{5}{21} \approx 0.24.
\end{equation*}
The judge rescales each fraction to the common $0$--$9$ scale (Section~\ref{sec:impl:metrics}) by multiplying by $9$, yielding the reported scores in Table~\ref{tab:worked-scores}.

\begin{table}[h]
\centering
\caption{Judge scores for test case $\tau_0$ (0--9 scale). The factoid-level metrics ($A$, $P$, $R$) are the focus of this paper; the remaining dimensions are the rubric-guided qualitative scores.}
\label{tab:worked-scores}
\small
\begin{tabular}{lc}
\toprule
\textbf{Dimension} & \textbf{Score} \\
\midrule
Precision      & 3.2 \\
Recall         & 3.8 \\
Accuracy       & 2.1 \\
Coverage       & 6.0 \\
Conciseness    & 7.5 \\
Tone           & 8.5 \\
Clarity        & 8.0 \\
\midrule
\textbf{Overall} & \textbf{5.1} \\
\bottomrule
\end{tabular}
\end{table}

\subsection{Why the scores}
Precision is low because several response factoids are incorrect or unsupported: the base Financial dataset IDs instead of the required Financial 6M IDs, a utilization-based target instead of the \texttt{billing.paymentFailed} mechanism, and a \texttt{transactionType='missed\_payment'} value that does not exist in the schema.
Recall is low because critical ground-truth factoids are absent from the response: the correct 6M dataset IDs, the \texttt{billing.paymentFailed} target event, the \texttt{count\_90d} event aggregations, the explicit $90$-day/$30$-day windows, and the \texttt{lastPaymentFailedAt} leakage exclusion.
This mirrors the interpretation in Section~\ref{sec:impl:metrics}: hallucinated or wrong facts depress precision, while omitted facts depress recall, and both are directly attributable in the scorecard (Table~\ref{tab:worked-issues}).

\begin{table}[h]
\centering
\caption{Issue categories flagged by the judge for test case $\tau_0$.}
\label{tab:worked-issues}
\small
\begin{tabular}{ll}
\toprule
\textbf{Category} & \textbf{Description} \\
\midrule
Wrong datasets            & Base Financial selected, not Financial 6M \\
Wrong target mechanism    & Utilization proxy, not \texttt{billing.paymentFailed} \\
Missing event aggregations & No \texttt{count\_90d} event features \\
Leakage not flagged       & \texttt{lastPaymentFailedAt} not excluded \\
Wrong observation window  & $90$d/$30$d windowing not defined \\
\bottomrule
\end{tabular}
\end{table}

\end{document}